\documentclass{article}
\usepackage[english]{babel}
\usepackage[utf8]{inputenc}
\usepackage{johd}
\usepackage{url,hyperref,lineno,microtype,subcaption}
\usepackage[onehalfspacing]{setspace}
\usepackage{booktabs}
\usepackage{graphicx}
\usepackage{amssymb}
\usepackage{amsmath}
\usepackage[table,xcdraw]{xcolor}
\usepackage[euler]{textgreek}

\title{Decoding EEG Brain Activity for Multi-Modal Natural Language Processing}

\author{Nora Hollenstein\,$^{1,*}$, Cedric Renggli\,$^{2}$, Benjamin Glaus\,$^{2}$, Maria Barrett\,$^{3}$,\\Marius Troendle\,$^{4}$, Nicolas Langer\,$^{4}$ and Ce Zhang\,$^{2}$\\
\small $^{1}$Department of Nordic Studies and Linguistics, University of Copenhagen \\
        \small $^{2}$Department of Computer Science, ETH Zurich, Switzerland \\
        \small $^{3}$Department of Computer Science, IT University of Copenhagen, Denmark \\
        \small $^{4}$Department of Psychology, University of Zurich, Switzerland \\\\
        \small $^{*}$Corresponding author: Nora Hollenstein; \tt{nora.hollenstein@hum.ku.dk}
}

\date{} %leave blank

\begin{document}

\maketitle

\begin{abstract} 
\noindent Until recently, human behavioral data from reading has mainly been of interest to researchers to understand human cognition. However, these human language processing signals can also be beneficial in machine learning-based natural language processing tasks. Using EEG brain activity for this purpose
is largely unexplored as of yet.
In this paper, we present the first large-scale study of systematically analyzing the potential of EEG brain activity data for improving natural language processing tasks, with a special focus on which features of the signal are most beneficial. We present a multi-modal machine learning architecture that learns jointly from textual input as well as from EEG features. We find that filtering the EEG signals into frequency bands is more beneficial than using the broadband signal. Moreover, for a range of word embedding types, EEG data improves binary and ternary sentiment classification and outperforms multiple baselines. For more complex tasks such as relation detection, only the contextualized BERT embeddings outperform the baselines in our experiments, which raises the need for further research. Finally, EEG data shows to be particularly promising when limited training data is available.  
\end{abstract}

\noindent\\

%\noindent\authorroles{For determining author roles, please use following taxonomy: \url{https://casrai.org/credit/}. Please list the roles for each author.} 

\section{Introduction}

\noindent Recordings of brain activity play an important role in furthering our understanding of how human language works \citep{murphy2018decoding,ling2019visual}. 
The appeal and added value of using brain activity signals in linguistic research are intelligible \citep{stemmer2012eeg}. 
Computational language processing models still struggle with basic linguistic phenomena that humans perform effortlessly \citep{ettinger2019bert}. Combining insights from neuroscience and artificial intelligence will take us closer to human-level language understanding \citep{mcclelland2020placing}. Moreover, numerous datasets of cognitive processing signals in naturalistic experiment paradigms with real-world language understanding tasks are becoming available \citep{kandylaki2019story,alday2019m}. 

\citet{linzen2020can} advocates for the grounding of NLP models in multi-modal settings to compare the generalization abilities of the models to human language learning. Multi-modal learning in machine learning refers to algorithms learning from multiple input modalities encompassing various aspects of communication.
Developing models that learn from such multi-modal inputs efficiently is crucial to advance the generalization capabilities of state-of-the-art NLP models. \citet{bisk2020experience} posit that text-only training seems to be reaching the point of diminishing returns and the next step in the development of NLP is leveraging multi-modal sources of information. Leveraging electroencephalography (EEG) and other physiological and behavioral signals seem especially appealing to model multi-modal human-like learning processes. Although combining different modalities or types of information for improving performance seems intuitively appealing, in practice, it is challenging to combine the varying level of noise and conflicts between modalities \citep{morency2017multimodal}. Therefore, we investigate if and how we take advantage of electrical brain activity signals to provide a human inductive bias for these natural language processing (NLP) models. 
%Our objective is to narrow the gap between human and machine language understanding.

Two popular NLP tasks are sentiment analysis and relation detection. The goal of both tasks is to automatically extract information from text. Sentiment analysis is the task of identifying and categorizing subjective information in text. For example, the sentence ``This movie is great fun." contains a positive sentiment, while the sentence ``This movie is terribly boring." contains a negative sentiment. Relation detection is the task of identifying semantic relationships between entities in the text. In the sentence ``Albert Einstein was born in Ulm.", the relation \textit{Birth Place} holds between the entities ``Albert Einstein" and ``Ulm". NLP researchers have made great progress in building computational models for these tasks \citep{barnes2017assessing,rotsztejn2018eth}. However, these machine learning (ML) models still lack core human language understanding skills that humans perform effortlessly \citep{poria-etal-2020-iceberg, barnes_velldal_ovrelid_2020}. \citet{barnes-etal-2019-sentiment} find that sentiment models struggle with different linguistic elements such as negations or sentences containing mixed sentiment towards several target aspects.

\subsection*{Leveraging Physiological Data for Natural Language Processing}

% encoding & decoding cognitive signals
\noindent The increasing wealth of literature on the cognitive neuroscience of language (see reviews by \citealp{poeppel2014neuroanatomic,poeppel2012towards,friederici2000developmental}) enables the use of cognitive signals in applied fields of language processing (e.g., \citealp{armeni2017probabilistic}). In recent years, natural language processing researchers have increasingly leveraged human language processing signals from physiological and neuroimaging recordings for both augmenting and evaluating machine learning-based NLP models (e.g., \citealp{artemova2020data,hollenstein2019cognival,toneva2019interpreting}). The approaches taken in those studies can be categorized as encoding or decoding cognitive processing signals.
Encoding and decoding are complementary operations: encoding uses stimuli to predict brain activity, while decoding uses the brain activity to predict information about the stimuli \citep{naselaris2011encoding}. In the present study, we focus on the decoding process for predicting information about the text input from human brain activity.

% related work in decoding, with ET and fMRI
Until now, mostly eye tracking and functional magnetic resonance imaging (fMRI) signals have been leveraged for this purpose (e.g., \citealp{fyshe2014interpretable}). On the one hand, fMRI recordings provide insights into the brain activity with a high spatial resolution, which furthers the research of localization of language-related cognitive processes. FMRI features are most often extracted over full sentences or longer text spans, since the extraction of word-level signals is highly complex due to the lower temporal resolution and hemodynamic delay. The number of cognitive processes and noise included in brain activity signals make feature engineering challenging. Machine learning studies leveraging brain activity data rely on standard preprocessing steps such as motion correction and spatial smoothing, and then use data-driven approaches to reduce the number of features, e.g., principal component analysis \citep{beinborn2019robust}. \citet{schwartz2019inducing}, for instance, fine-tuned a contextualized language model with brain activity data, which yields better predictions of brain activity and does not harm the model's performance on downstream NLP tasks. On the other hand, eye tracking enables us to objectively and accurately record visual behavior with high temporal resolution at low cost. Eye tracking is widely used in psycholinguistic studies and it is common to extract well-established theory-driven features \citep{hollenstein2020towards,mathias2020survey,barrett2016weakly}. These established metrics are derived from a large body of psycholinguistic research.

% introduce EEG here
EEG is a non-invasive method to measure electrical brain surface activity. The synchronized activity of neurons in the brain produces electrical currents. The resulting voltage fluctuations can be recorded with external electrodes on the scalp. Compared to fMRI and other neuroimaging techniques, EEG can be recorded with a very high temporal resolution. This allows for more fine-grained language understanding experiments on the word-level, which is crucial for applications in NLP \citep{beres2017time}. To isolate certain cognitive functions, EEG signals can be split into frequency bands. For instance, effects related to semantic violations can be found within the gamma frequency range ($\sim30-100$ Hz), with well-formed sentences showing higher gamma levels than sentences containing violations \citep{penolazzi2009gamma}. Due to the wide extent of cognitive processes and the low signal-to-noise ratio in the EEG data, it is very challenging to isolate specific cognitive processes, so that more and more researchers are relying on machine learning techniques to decode the EEG signals \citep{sun2019brain2char,affolter2020brain2word,pfeiffer2020neural}. These challenges are the decisive factors why EEG has not yet been used for NLP tasks. Data-driven approaches combined with the possibility of naturalistic reading experiments are now bypassing these challenges.

% why EEG might be good for NLP
Reading times of words in a sentence depend on the amount of information the words convey. This correlation can be observed in eye tracking data, but also in EEG data \citep{frank2015erp}. Thus, eye tracking and EEG are complementary measures of cognitive load. Compared to eye tracking, EEG may be more cumbersome to record and requires more expertise. Nevertheless, while eye movements indirectly reflect the cognitive load of text processing, EEG contains more direct and comprehensive information about language processing in the human brain. As we show below, this is beneficial for the higher level semantic NLP tasks targeted in this work.
For instance, word predictability and semantic similarity show distinct patterns of brain activity during language comprehension \citep{frank2017wordpred,ettinger2019bert}. The word representations used by neural networks and brain activity observed via the process of subjects reading a story can be aligned \citep{wehbe2014aligning}. Moreover, EEG effects that reflect syntactical processes can also be found in computational models of grammar \citep{hale2018finding}.

% add why this is important for NLP
The co-registration of EEG and eye-tracking has become an important tool for studying the temporal dynamics of naturalistic reading \citep{dimigen2011coregistration,sato2018successful,hollenstein2018zuco}. This methodology has been increasingly and successfully used to study EEG correlates in the time domain (i.e., event-related potentials, ERPs) of cognitive processing in free viewing situations such as reading \citep{degno2021co}. In this context, fixation-related potentials (FRPs), which are the evoked electrical responses time-locked to the onset of fixations, have been studied and have received broad interest by naturalistic imaging researchers for free viewing studies. In naturalistic reading paradigms, FRPs allow the study of the neural dynamics of how novel information from currently fixated text affects the ongoing language comprehension process.

%In NLP, EEG has only been used rarely until now. 
As of yet, the related work relying on EEG signals for NLP is very limited. \citet{sassenhagen2020traces} find that word embeddings can successfully predict the pattern of neural activation. However, their experiment design does not include natural reading, but reading isolated words. \citet{hollenstein2019cognival} similarly find that various embedding type are able to predict aggregate word-level activations from natural reading, where contextualized embeddings perform best. Moreover, \citet{murphy2010detecting} showed that semantic categories can be detected in simultaneous EEG recordings. \citet{muttenthaler2020human} used EEG signals to train an attention mechanism, similar to \citet{barrett2018sequence}, who used eye tracking signals to induce machine attention with human attention. However, EEG has not yet been leveraged for higher-level semantic tasks such as sentiment analysis or relation detection. Deep learning techniques have been applied to decode EEG signals \citep{craik2019deep}, especially for brain-computer interface technologies, e.g., \citet{nurse2016decoding}. However, this avenue has not yet been explored when leveraging EEG signals to enhance NLP models. Through decoding EEG signals occurring during language understanding, more specifically, during English sentence comprehension, we aim to explore their impact on computational language understanding tasks.

\subsection*{Contributions}
\noindent More than a practical application of improving real-world NLP tasks, our main goal is to explore to what extent there is additional linguistic processing information in the EEG signal to complement the text input. In this present study, we investigate for the first time the potential of leveraging EEG signals for augmenting NLP models. For the purpose of making language decoding studies from brain activity more interpretable, we follow the recommendations of \citet{gauthier2018does}: (1) We commit to a specific mechanism and task, and (2) subdivide the input feature space including theoretically founded preprocessing steps. We investigate the impact of enhancing a neural network architecture for two common NLP tasks with a range of EEG features. We propose a multi-modal network capable of processing textual features and brain activity features simultaneously. We employ two different well-established types of neural network architectures for decoding the EEG signals throughout the entire study. To analyze the impact of different EEG features, we perform experiments on sentiment analysis as a binary or ternary sentence classification task, and relation detection as a multi-class and multi-label classification task. We investigate the effect of augmenting NLP models with neurophysiolgical data in an extensive study while accounting for various dimensions:

\begin{enumerate}
  \item We present a comparison of a purely data-driven approach of feature extraction for machine learning, using full \textit{broadband EEG signals}, to a more theoretically motivated approach, splitting the word-level EEG features into \textit{frequency bands}.
  \item  We develop two \textit{EEG decoding components} for our multi-modal ML architecture: A recurrent and a convolutional component.
  \item We contrast the effects of these EEG features on multiple \textit{word representation types} commonly used in NLP. We compare the improvements of EEG features as a function of various \textit{training data sizes}.
  \item We analyze the impact of the EEG features on varying \textit{classification complexity}: from binary classification to multi-class and multi-label tasks.
\end{enumerate}

\noindent This comprehensive study is completed by comparing the impact of the decoded EEG signals not only to a \textit{text-only} baseline, but also to baselines augmented with eye tracking data as well as random noise. In the next section, we describe the materials used in this study and the multi-modal machine learning architecture. Thereafter, we present the results of the NLP tasks and discuss the dimensions defined above.

\section{Materials and Methods}

\subsection{Data}

\noindent For the purpose of augmenting natural language processing tasks with brain activity signals, we leverage the Zurich Cognitive Language Processing Corpus (ZuCo; \citealp{hollenstein2018zuco,hollenstein2020zuco}). ZuCo is an openly available dataset of simultaneous EEG and eye tracking data from subjects reading naturally occurring English sentences. This corpus consists of two datasets, ZuCo 1.0 and ZuCo 2.0, which contain the same type of recordings. We select the normal reading paradigms from both datasets, in which participants were instructed to read English sentences in their own pace with no specific task beyond reading comprehension. The participants read one sentence at a time, using a control
pad to move to the next sentence. This setup facilitated the naturalistic reading paradigm. Descriptive statistics about the datasets used in this work are presented in Table \ref{tab:dataset}.

\begin{table}[t]
\centering
%\small
\begin{tabular}{lccc}
\toprule
 & \textbf{ZuCo 1.0} & \textbf{ZuCo 1.0} & \textbf{ZuCo 2.0} \\
 & Task SR & Task NR & Task NR \\\midrule
Participants & 12 & 12 & 18 \\
Sentences & 400 & 300 & 349 \\
Words & 7,079 & 6,386 & 6,828\\
Unique word types & 3,080 & 2,657 & 2,412\\
\midrule
Sentiment Analysis & \checkmark & - & - \\
Relation Detection & - & \checkmark & \checkmark \\\bottomrule
\end{tabular}
\caption{Details about the ZuCo tasks used in this paper. In \textit{Task SR} participants read sentences from movie reviews, and in \textit{Task NR} sentences from Wikipedia articles.}
\label{tab:dataset}
\end{table}

A detailed description of the entire ZuCo dataset, including individual reading speed, lexical performance, average word length, average number of words per sentence, skipping proportion on word level, and effect of word length on skipping proportion, can be found in \citet{hollenstein2018zuco}. In the following section, we will describe the methods relevant to the subset of the ZuCo data used in the present study.

\subsubsection{Participants}

\noindent For ZuCo 1.0, data were recorded from 12 healthy adults (between 22 and 54 years old; all right-handed; 5 female subjects).
For ZuCo 2.0, data were recorded from 18 healthy adults (between 23 and 52 years old; 2 left-handed; 10 female subjects). 
The native language of all participants is English, originating from Australia, Canada, UK, USA or South Africa. In addition, all subjects completed the standardized LexTALE test to assess their vocabulary and language proficiency (Lexical Test for Advanced Learners of English; \citealp{lemhofer2012introducing}). All participants gave written consent for their participation and the re-use of the data prior to the start of the experiments. The study was approved by the Ethics Commission of the University of Zurich.

\subsubsection{Reading Materials \& Experimental Design}

The reading materials recorded for the ZuCo corpus contain sentences from movie reviews from the Stanford Sentiment Treebank \citep{socher2013recursive} and Wikipedia articles from a dataset provided by \citet{culotta2006integrating}. These resources were chosen since they provide ground truth labels for the machine learning tasks in this work. Table \ref{tab:task-examples} presents a few examples of the sentences read during the experiments.

For the recording sessions, the sentences were presented one at a time at the same position on the screen. Text was presented in black with font size 20-point Arial on a light grey background resulting in a letter height of 0.8 mm or 0.674°. The lines were triple-spaced, and the words double-spaced. A maximum of 80 letters or 13 words were presented per line in all three tasks. Long sentences spanned multiple lines. A maximum of 7 lines for Task 1, 5 lines for Task 2 and 7 lines for Task 3 were presented simultaneously on the screen.

\noindent During the normal reading tasks included in the ZuCo corpus, the participants were instructed to read the sentences naturally, without any specific task other than comprehension. Participants were told to read the sentences normally without any special instructions. Participants were equipped with a control to trigger the onset of the next sentence. The task was explained to the subjects orally, followed by instructions on the screen.

The control condition for this task consisted of single-choice reading comprehension questions about the content of the previous sentence. 12\% of randomly selected sentences were followed by a control question on a separate screen. To test the participants' reading comprehension, these questions ask about the content of the previous sentence. The questions are presented with three answer options, out of which only one is correct.

\subsubsection{EEG Data}

\noindent In this section, we present the EEG data extracted from the ZuCo corpus for this work. We describe the acquisition and preprocessing procedures as well as the feature extraction. 

\paragraph{EEG Acquisition and Preprocessing}
% EEG acquisition and preprocessing
\noindent High-density EEG data were recorded using a 128-channel EEG Geodesic Hydrocel system (Electrical Geodesics, Eugene, Oregon)  with a sampling rate of 500 Hz. The recording reference was at Cz (vertex of the head), and the impedances were kept below 40 kΩ. All analyses were performed using MATLAB 2018b (The  MathWorks, Inc.,  Natick, Massachusetts, United States). EEG data was automatically preprocessed using the current version (2.4.3) of \textit{Automagic} \citep{pedroni2019automagic}. Automagic is an open-source MATLAB toolbox that acts as a wrapper to run currently available EEG preprocessing methods and offers objective standardized quality assessment for large studies. The code for the preprocessing can be found online.\footnote{\url{https://github.com/methlabUZH/automagic}}

Our preprocessing pipeline consisted of the following steps. First, 13 of the 128 electrodes in the outermost circumference (chin and neck) were excluded from further processing as they capture little brain activity and mainly record muscular activity. Additionally, 10 EOG electrodes were used for blink and eye movement detection (and subsequent rejection) during ICA. The EOG electrodes were removed from the data after the preprocessing, yielding a total number of 105 EEG electrodes. Subsequently, \textit{bad} channels were detected by the algorithms implemented in the EEGLAB plugin \texttt{clean\_rawdata}\footnote{\url{http://sccn.ucsd.edu/wiki/Plugin\_list\_process}}, which removes flatline, low-frequency, and noisy channels. A channel was defined as a bad electrode when recorded data from that electrode was correlated at less than 0.85 to an estimate based on neighboring channels. Furthermore, a channel was defined as bad if it had more line noise relative to its signal than all other channels (4 standard deviations). Finally, if a channel had a longer flat-line than 5 seconds, it was considered bad. These bad channels were automatically removed and later interpolated using a spherical spline interpolation (EEGLAB function \texttt{eeg\_interp.m}). The interpolation was performed as a final step before the automatic quality assessment of the EEG files (see below). Next, data was filtered using a high-pass filter (-6dB cut-off: 0.5 Hz) and a 60 Hz notch filter was applied to remove line noise artifacts. Thereafter, an independent component analysis (ICA) was performed. Components reflecting artifactual activity were classified by the pre-trained classifier MARA \citep{winkler2011automatic}. MARA is a supervised machine learning algorithm that learns from expert ratings. Therefore, MARA is not limited to a specific type of artifact, and should be able to handle eye artifacts, muscular artifacts and loose electrodes equally well. Each component being classified with a probability rating $>0.5$ for any class of artifacts was removed from the data. Finally, residual bad channels were excluded if their standard deviation exceeded a threshold of 25 μV. After this, the pipeline automatically assessed the quality of the resulting EEG files based on four criteria: A data file was marked as bad-quality EEG and not included in the analysis if (1) the proportion of high-amplitude data points in the signals ($>30$ μV) was larger than 0.20; (2) more than 20\% of time points showed a variance larger than 15 microvolts across channels; (3) 30\% of the channels showed high variance ($>15$ μV); (4) the ratio of bad channels was higher than 0.3. 

Free viewing in reading is an important characteristic of naturalistic behavior and imposes challenges for the analysis of electrical brain activity data. Free viewing in the context of our study refers to the participant’s ability to perform self-paced reading given the experimental requirement to keep the head still during data recording. In the case of EEG recordings during naturalistic reading, the self-paced timing of eye fixations leads to a temporal overlap between successive fixation-related events \citep{dimigen2011coregistration}. In order to isolate the signals of interest and correct for temporal overlap in the continuous EEG, several methods using linear-regression-based deconvolution modeling have been proposed for estimating the overlap-corrected underlying neural responses to events of different types (e.g., \citealt{ehinger2019unfold,smith2015regression,smith2015regression2}). Here, we used the \textit{unfold} toolbox for MATLAB \citep{ehinger2019unfold}.\footnote{\url{https://github.com/unfoldtoolbox/unfold/}} Deconvolution modeling is based on the assumption that in each channel the recorded signal consists of a combination of time-varying and partially overlapping event-related responses and random noise. Thus, the model estimates the latent event-related responses to each type of event based on repeated occurrences of the event over time, in our case eye fixations.

\paragraph{EEG Features}

\noindent The fact that ZuCo provides simultaneous EEG and eye tracking data highly facilitates the extraction of word-level brain activity signals. \citet{dimigen2011coregistration} demonstrated that EEG indices of semantic processing can be obtained in natural reading and compared to eye movement behavior. The eye tracking data provides millisecond-accurate fixation times for each word. Therefore, we were able to obtain the brain activity during all fixations of a word by computing fixation-related potentials aligned to the onsets of the fixation on a given word.

\begin{table}[t]
\centering
\begin{tabular}{lll}
\toprule
\textbf{Task} & \textbf{Example Sentence} & \textbf{Label(s)} \\\midrule
Binary/ternary sentiment analysis & \begin{tabular}[c]{@{}l@{}}``The film often achieves a mesmerizing poetry."\end{tabular} & \textit{Positive} \\\midrule
Binary/ternary sentiment analysis & \begin{tabular}[c]{@{}l@{}}``Flaccid drama and exasperatingly slow journey."\end{tabular} & \textit{Negative} \\\midrule
Ternary sentiment analysis & ``A portrait of an artist." & \textit{Neutral} \\\midrule
Relation detection & ``He attended Wake Forest University." & \textit{Education} \\\midrule
Relation detection & \begin{tabular}[c]{@{}l@{}}``She attended Beverly Hills High School, but \\ left to become an actress."\\\end{tabular} & \begin{tabular}[c]{@{}l@{}}\textit{Education}, \\ \textit{Job Title}\\
\end{tabular}
\\\bottomrule
\end{tabular}
\caption{Example sentences for all three NLP tasks used in this study.}
\label{tab:task-examples}
\end{table}

% EEG features
In this work, we select a range of EEG features with a varying degree of theory-driven and data-driven feature extraction. We define the \textit{broadband EEG signal}, i.e., the full EEG signal from 0.5-50 Hz as the averaged brain activity over all fixations of a word, i.e., its total reading time. We compare the full EEG features, a data-driven feature extraction approach, to \textit{frequency band features}, a more theoretically motivated approach. Different neurocognitive aspects of language processing during reading are associated with brain oscillations at various frequencies. These frequency ranges are known to be associated with certain cognitive functions. We split the EEG signal into four frequency bands to limit the bandwidth of the EEG signals to be analyzed. The frequency bands are fixed ranges of wave frequencies and amplitudes over a time scale: \textit{theta} (4-8 Hz), \textit{alpha} (8.5-13 Hz), \textit{beta} (13.5-30 Hz), and \textit{gamma} (30.5-49.5 Hz). We elaborate on cognitive and linguistic functions of each of these frequency bands in Section \ref{sec:disc-features}.

We then applied a Hilbert transform to each of these time-series, resulting in a complex time-series. The Hilbert phase and amplitude estimation method yields results equivalent to sliding window FFT and wavelet approaches \citep{bruns2004fourier}. We specifically chose the Hilbert transformation to maintain temporal information for the amplitude of the frequency bands to enable the power of the different frequencies for time segments defined through fixations from the eye-tracking recording. Thus, for each eye-tracking feature we computed the corresponding EEG feature in each frequency band. For each EEG eye-tracking feature, all channels were subject to an artifact rejection criterion of $\pm90$ μV to exclude trials with transient noise.

In spite of the high inter-subject variability in EEG data, it has been shown in previous research of machine learning applications \citep{foster2018decoding,hollenstein2019advancing}, that averaging over the EEG features of all subjects yields results almost as good as the single best-performing subjects. Hence, we also average the EEG features over all subjects to obtain more robust features. Finally, for each word in each sentence, the EEG features consist of a vector of 105 dimensions (one value for each EEG channel). For training the ML models, we split all available sentences into sets of 80\% for training and 20\% for testing to ensure that the test data is unseen during training.\\

\subsubsection{Eye Tracking Data}\label{sec:et-preprocessing}

\noindent In the following, we describe the eye tracking data recorded for the Zurich Cognitive Language Processing Corpus. In this study, we focus on decoding EEG data, but we use eye movement data to compute an additional baseline. As mentioned previously, augmenting ML models with eye tracking yields consistent improvements across a range of NLP tasks, including sentiment analysis and relation extraction \citep{long2017cognition,mishra2017leveraging,hollenstein2019advancing}. Since the ZuCo datasets provide simultaneous EEG and eye tracking recordings, we leverage the available eye tracking data to augment all NLP tasks with eye tracking features as an additional multi-modal baseline based on cognitive processing features.

\paragraph{Eye Tracking Acquisition and Preprocessing} 

\noindent Eye movements were recorded with an infrared video-based eye tracker (EyeLink 1000 Plus, SR Research) at a sampling rate of 500 Hz. The EyeLink 1000 tracker processes eye position data, identifying saccades, fixations and blinks. Fixations were defined as time periods without saccades during which the visual gaze is fixed on a single location. The data therefore consists of (x,y) gaze location entries for individual fixations mapped to word boundaries. A fixation lasts around 200--250ms (with large variations). Fixations shorter than 100 ms were excluded, since these are unlikely to reflect language processing \citep{sereno2003measuring}. Fixation duration depends on various linguistic effects, such as word frequency, word familiarity and syntactic category \citep{clifton2007eye}.

\paragraph{Eye Tracking Features}\label{sec:et_features}

\noindent The following features were extracted from the raw data: (1) \textit{gaze duration} (GD), the sum of all fixations on the current word in the first-pass reading before the eye moves out of the word; (2) \textit{total reading time} (TRT), the sum of all fixation durations on the current word, including regressions; (3) \textit{first fixation duration} (FFD), the duration of the first fixation on the prevailing word; (4) \textit{go-past time} (GPT), the sum of all fixations prior to progressing to the right of the current word, including regressions to previous words that originated from the current word; (5) number of fixations (nFix), the total amount of fixations on the current word.

We use these five features provided in the ZuCo dataset, which cover the extent of the human reading process. To increase the robustness of the signal, analogously to the EEG features, the eye tracking features are averaged over all subjects \citep{barrett2020sequence}. This results in a feature vector of five dimensions for each word in a sentence. Training and test data were split in the same fashion as the EEG data.

\subsection{Natural Language Processing Tasks}

\noindent In this section, we describe the natural language processing tasks we use to evaluate the multi-modal ML models. As usual in supervised machine learning, the goal is to learn a mapping from given input features to an output space to predict the labels as accurately as possible. 
The tasks we consider in our work do not differ much in the input definition as they consist of three sequence classification tasks for information extraction from text. The goal of a sequence classification task is to assign the correct label(s) to a given sentence. The input for all tasks consists of tokenized sentences, which we augment with additional features, i.e., EEG or eye tracking. The labels to predict vary across the three chosen tasks resulting in varying task difficulty. 
Table \ref{tab:task-examples} provides examples for all three tasks.

\subsubsection{Task 1 and 2: Sentiment Analysis} 
\noindent The objective of sentiment analysis is to interpret subjective information in text. More specifically, we define sentiment analysis as a sentence-level classification task. We run our experiments on both binary (\textit{positive/negative}) and ternary (\textit{+ neutral}) sentiment classification. For this task, we leverage only the sentences recorded in the first task of ZuCo 1.0, since they are part of the Stanford Sentiment Treebank \citep{socher2013recursive}, and thus directly provide annotated sentiment labels for training the ML models. 
For the first task, binary sentiment analysis, we use the 263 positive and negative sentences. For the second task, ternary sentiment analysis, we additionally use the neutral sentences, resulting in a total of 400 sentences.

\subsubsection{Task 3: Relation Detection} 
\begin{figure}[t]
	\centering
 \includegraphics[width=0.95\textwidth]{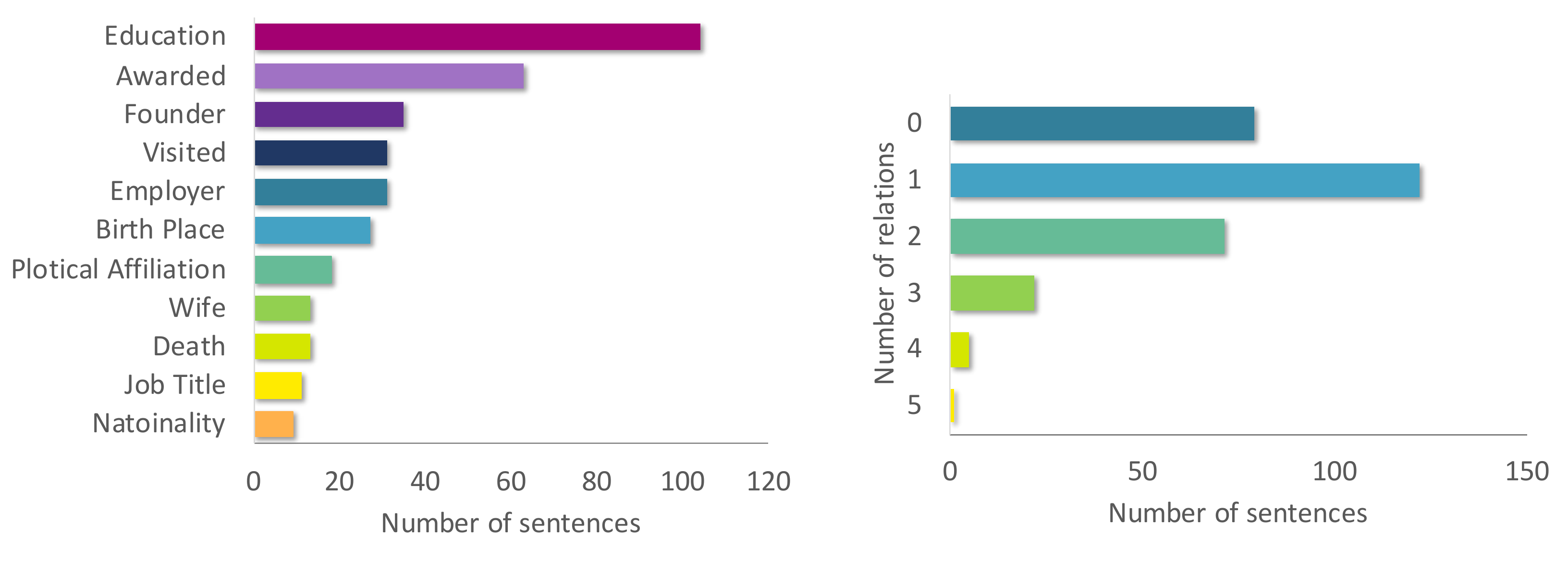}
	\caption{(left) Label distribution of the 11 relation types in the relation detection dataset. (right) Number of relation types per sentence in the relation detection dataset.}
	\label{fig:deco-rel-distr}
\end{figure}

\noindent Relation classification is the task of identifying the semantic relation holding between two entities in text. The ZuCo corpus also contains Wikipedia sentences with relation types such as \textit{Job Title}, \textit{Nationality} and \textit{Political Affiliation}. The sentences in ZuCo 1.0 and ZuCo 2.0, from the normal reading experiment paradigms, include 11 relation types (Figure \ref{fig:deco-rel-distr}). In order to further increase the task complexity, we treat this task differently than the sentiment analysis tasks. Since any sentence can include zero, one or more of the relevant semantic relations (see example in Table \ref{tab:task-examples}, we treat relation detection as a multi-class \textit{and} multi-label sequence classification task. Concretely, every sample can be assigned to any possible combination out of the 11 classes including none of them. Removing duplicates between ZuCo 1.0 and ZuCo 2.0 resulted in 594 sentences used for training the models. Figure \ref{fig:deco-rel-distr} illustrates the label and relation distribution of the sentences used to train the relation detection task.

\subsection{Multi-Modal Machine Learning Architecture}

\noindent We present a multi-modal neural architecture to augment the NLP sequence classification tasks with any other type of data. Although combining different modalities or types of information for improving performance seems an intuitively appealing task, it is often challenging to combine the varying levels of noise and conflicts between modalities in practice. 

Previous works using physiological data for improving NLP tasks mostly implement \textit{early fusion} multi-modal methods, i.e., directly concatenating the textual and cognitive embeddings before inputting them into the network. For example, \citet{hollenstein2019entity}, \citet{barrett2018unsupervised} and \citet{mishra2017leveraging} concatenate textual input features with eye-tracking features to improve NLP tasks such as entity recognition, part-of-speech tagging and sentiment analysis, respectively. Concatenating the input features at the beginning in only one joint decoder component aims at learning a joint decoder across all modalities at risk of implicitly learning different weights for each modality. However, recent multi-modal machine learning work has shown the benefits of \textit{late fusion} mechanisms \citep{ramachandram2017deep}. \citet{do2017multiview} argument in favor of concatenating the hidden layers instead of concatenating the features at input time. Such multi-modal models have been successfully applied in other areas, mostly combining inputs across different domains, for instance, learning speech reconstruction from silent videos \citep{ephrat2017improved}, or for text classification using images \citep{kiela2018efficient}. \citet{tsai2019multimodal} train a multi-modal sentiment analysis model from natural language, facial gestures, and acoustic behaviors. 

Hence, we adopted the late fusion strategy in our work. We present multi-modal models for various NLP tasks, combining the learned representations of all input types (i.e., text and EEG features) in a late fusion mechanism before conducting the final classification. Purposefully, this enables the model to learn independent decoders for each modality before fusing the hidden representations together. In the present study, we investigate the proposed multi-modal machine learning architecture, which learns simultaneously from text and from cognitive data such as eye tracking and EEG signals.

In the following, we first describe the uni-modal and multi-modal baseline models we use to evaluate the results. Thereafter, we present the multi-modal NLP models that jointly learn from text and brain activity data.

\begin{figure}[t]
	\centering
 \includegraphics[width=1\textwidth]{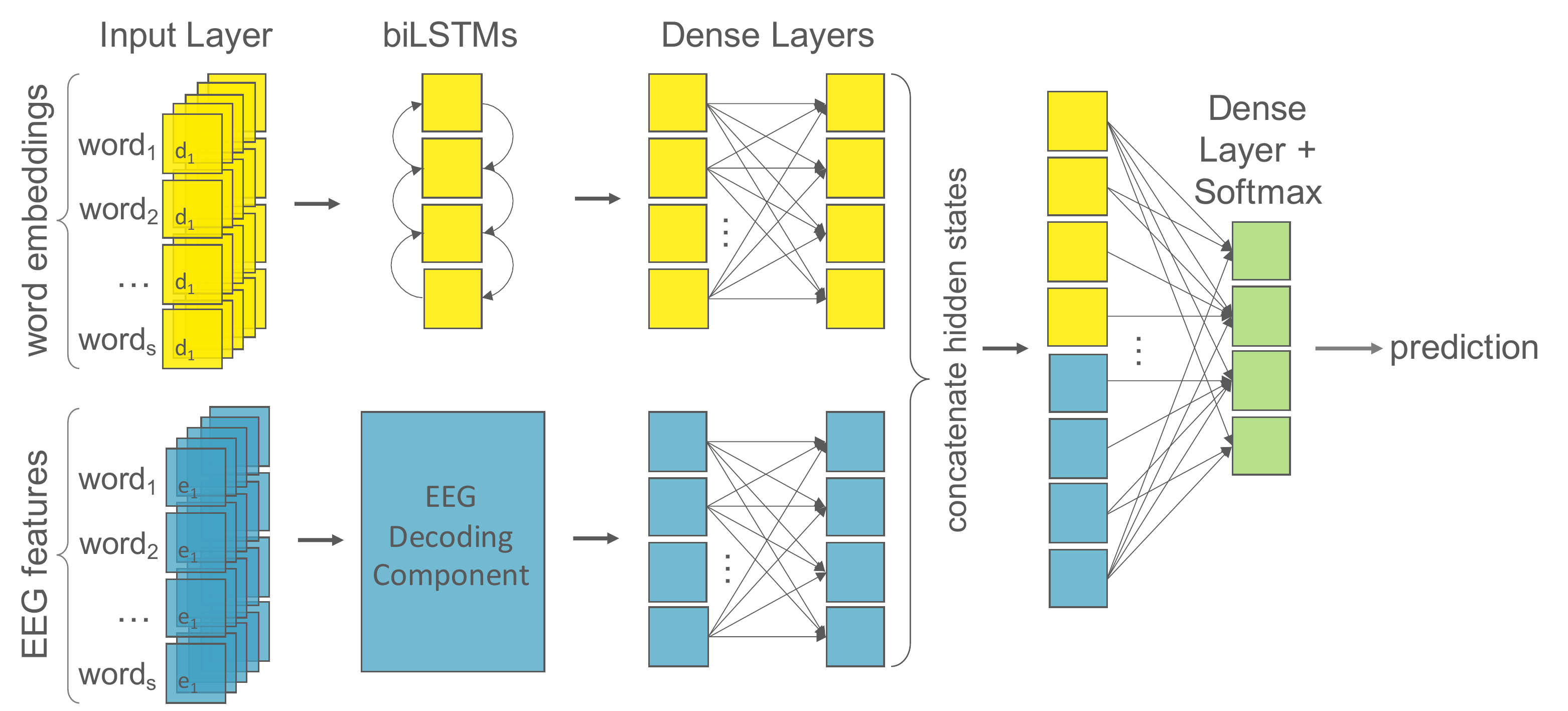}
	\caption{The multi-modal machine learning architecture for the EEG-augmented models. Word embeddings of dimension $d$ are the input for the textual component (yellow); EEG features of dimension $e$ for the cognitive component (blue). The text component consists of recurrent layers followed by two dense layers with dropout. We test multiple architectures for the EEG component (see Figure \ref{fig:eeg-components}). Finally, the hidden states of both components are concatenated and followed by a final dense layer with softmax activation for classification (green).}
	\label{fig:architecture}
\end{figure}

\subsubsection{Uni-Modal Text Baselines}\label{sec:text-base}

\noindent For each of the tasks presented above, we train uni-modal models on textual features only. To represent the word numerically, we use word embeddings. Word embeddings are vector representations of words, computed so that words with similar meaning have a similar representation. To analyze the interplay between various types of word embeddings and EEG data, we use the following three embedding types typically used in practice: (1) randomly initialized embeddings trained at run time on the sentences provided, (2) GloVe pre-trained embeddings based on word co-occurrence statistics \citep{pennington2014glove}\footnote{\url{https://nlp.stanford.edu/projects/glove/}}, and (3) BERT pre-trained contextual embeddings \citep{devlin2019bert}\footnote{\url{https://huggingface.co/bert-base-uncased}}. 

The randomly initialized word representations define word embeddings as \textit{n}-by-\textit{d} matrices, where \textit{n} is the vocabulary size, i.e., the number of unique words in our dataset, and \textit{d} is the embedding dimension. Each value in that matrix is randomly initialized and will then be trained together with the neural network parameters. We set $d = 32$. This type of embeddings does not benefit from pre-training on large text collections and hence is known to perform worse than GloVe or BERT embeddings. We include them in our study to better isolate the impact of the EEG features and to limit the learning of the model on the text it is trained on.
Non-contextual word embeddings such as GloVe encode each word in a fixed vocabulary as a vector. The purpose of these vectors is to encode semantic information about a word, such that similar words result in similar embedding vectors. We use the GloVe embeddings of $d=300$ dimensions that are trained on 6 billion words.
The contextualized BERT embeddings were pre-trained on multiple layers of transformer models with self-attention \citep{vaswani2017attention}. Given a sentence, BERT encodes each word into a feature vector of dimension $d = 768$, which incorporates information from the word's context in the sentence.

The uni-modal text baseline model consists of a first layer taking the embeddings as an input, followed by a bidirectional Long-Short Term Memory network (LSTM; \citealp{hochreiter1997long}), then two fully-connected dense layers with dropout between them, and finally a prediction layer using softmax activation. This corresponds to a single component of the multi-modal architecture, i.e., the top component in Figure \ref{fig:architecture}. Following best practices~\citep[e.g.,][]{sun2019fine} , we set the weights of BERT to be trainable similarly to the randomly initialized embeddings. This process of adjusting the initialized weights of a pre-trained feature extractor during the training process, in our case BERT, is commonly known as \textit{fine-tuning} in the literature~\citep{howard2018universal}. In contrast, the parameters of the GloVe embeddings are fixed to the pre-trained weights and thus do not change during training.

\subsubsection{Multi-Modal Baselines}\label{sec:aug-base}

\noindent To analyze the effectiveness of our multi-modal architecture with EEG signals properly, we not only compare it to uni-modal text baselines, but also to multi-modal baselines using the same architecture described in the next section for the EEG models, but replacing the features of the second modality with the following alternatives: (1) We implement a gaze-augmented baseline, where the five eye tracking features described in Section \ref{sec:et_features} are combined with the word embeddings by adding them to the multi-modal model in the same manner as the EEG features, as vectors with dimension = 5. The purpose of this baseline is to allow a comparison of multi-modal models learning from two different types of physiological features. Since the benefits of eye tracking data in ML models are well established \citep{barrett2020sequence,mathias2020survey}, this is a strong baseline. (2) We further implement a random noise-augmented baseline, where we add uniformly sampled vectors of random numbers as the second input data type to the multi-modal model. These random vectors are of the same dimension as the EEG vectors (i.e., $d = 105$). It is well known that the addition of noise to the input data of a neural network during training can lead to improvements in generalization performance as a form of regularization \citep{bishop1995training}. Thus, this baseline is relevant because we want to analyze whether the improvements from the EEG signals on the NLP tasks are due to its capability of extracting linguistic information and not merely due to additional noise.

\subsubsection{EEG Models}\label{sec:eeg-models}

\begin{figure}[t]
	\centering
 \includegraphics[width=1\textwidth]{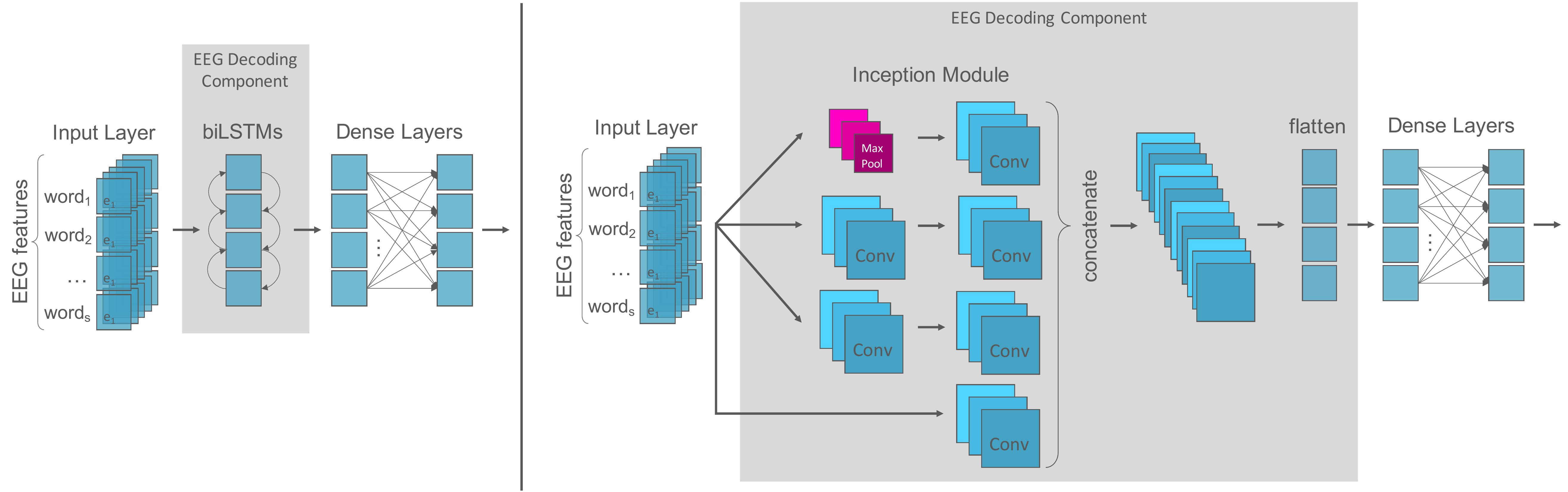}
	\caption{EEG decoding components: (left) The recurrent model component is analogous to the text component and consists of recurrent layers followed by two dense layers with dropout. (right) The convolutional inception component consists of an ensemble of convolution filters of varying lengths which are concatenated and flattened before the subsequent dense layers.}
	\label{fig:eeg-components}
\end{figure}

\noindent To fully understand the impact of the EEG data on the NLP models, we build a model that is able to deal with multiple inputs and mixed data. We present a multi-modal model with late decision-level fusion to learn joint representations of textual and cognitive input features. We test both a recurrent and a convolutional neural architecture for decoding the EEG signals.
Figure \ref{fig:architecture} depicts the main structure of our model and we describe the individual components below.

All input sentences are padded to the maximum sentence length to provide fixed-length text inputs to the model. Word embeddings of dimension $d$ are the input for the textual component, where $d \in \{32,300,768\}$ for randomly initialized embeddings, GloVe embeddings and BERT embeddings, respectively. EEG features of dimension $e$ are the input for the cognitive component, where $e=105$. As described, the text component consists of bidirectional LSTM layers followed by two dense layers with dropout. Text and EEG features are given as independent inputs to their own respective component of the network. The hidden representations of these are then concatenated before being fed to a final dense classification layer.We also experimented with different merging mechanisms to join the text and EEG layers of our two-tower model (concatenation, addition, subtraction, maximum). Concatenation overall achieved the best results, so we report only these. Although the goal of each network is to learn feature transformations for their own modality, the relevant extracted information should be complementary. This is achieved, as commonly done in deep learning, through alternatively running inference and back-propagation of the data through the entire network enabling information to \textit{flow} from the component responsible for one input modality to the other via the fully connected output layers. To learn a non-linear transformation function for each component, we employ the rectified linear units (ReLu) as activation functions after each hidden layer.

For the EEG component, we test a recurrent and a convolutional architecture since both have proven useful in learning features from time series data for language processing (e.g., \citealp{fawaz2020inceptiontime,yin2017comparative,lipton2015critical}). For the recurrent architecture (Figure \ref{fig:eeg-components}, left), the model component is analogous to the text component: it consists of bidirectional LSTM layers followed by two dense layers with dropout and ReLu activation functions. For the convolutional architecture (Figure \ref{fig:eeg-components}, right), we build a model component based on the \textit{Inception} module first introduced by \citet{szegedy2015going}. An \textit{inception} module is an ensemble of convolutions that applies multiple filters of varying lengths simultaneously to an input time series. This allows the network to automatically extract relevant features from both long and short time series. As suggested by \citet{schirrmeister2017deep} we used exponential linear unit activations (ELUs; \citealp{clevert2015fast}) in the convolutional EEG decoding model component.
 
For binary and ternary sentiment analysis, the final dense layer has a softmax activation in order to use the maximal output for the classification. For the multi-label classification case of relation detection, we replace the softmax function in the last dense layer of the model with a sigmoid activation to produce independent scores for each class. If the score for any class surpasses a certain threshold, the sentence is labeled to contain that relation type (opposite to simply taking the \textit{max} score as the label of the sentence). The threshold is tuned as an additional hyper-parameter.

This multi-modal model with separate components learned for each input data type has several advantages: It allows for separate pre-processing of each type of data. For instance, it is able to deal with differing tokenization strategies, which is useful in our case since it is challenging to map linguistic tokenization to the word boundaries presented to participants during the recordings of eye tracking and brain activity. Moreover, this approach is scalable to any number of input types. The generalizability of our model enables the integration of multiple data representations, e.g., learning from brain activity, eye movements, and other cognitive modalities simultaneously.

\subsubsection{Training Setup}

\begin{table}[t]
\centering
\small
\begin{tabular}{l|l}
\toprule
\textbf{Parameter} & \textbf{Range} \\\midrule
LSTM layer dimension & 64, 128, 256, 512 \\
Number of LSTM layers & 1, 2, 3, 4 \\\midrule
CNN filters & 14, 16, 18 \\
CNN kernel sizes & [1,4,7] \\
CNN pool sizes & 3, 5, 7 \\\midrule
Dense layer dimension & 8, 16, 32, 64, 128, 256, 512 \\
Dropout & 0.1, 0.3, 0.5 \\
Batch size & 20, 40, 60 \\
Learning rate & $10^{-1}$, $10^{-2}$, $10^{-3}$, $10^{-4}$, $10^{-5}$ \\
Random seeds & 13, 22, 42, 66, 78 \\\midrule
Threshold & 0.3, 0.5, 0.7 \\\bottomrule
\end{tabular}
\caption{Tested value ranges included in the hyper-parameter search for our multi-modal machine learning architecture. \textit{Threshold} only applies to relation detection.}
\label{tab:params}
\end{table}

\noindent To assess the impact of the EEG signals under fair modelling conditions, the hyper-parameters are tuned individually for all baseline models as well as for all eye tracking and EEG augmented models. The ranges of the hyper-parameters are presented in Table \ref{tab:params}. All results are reported as means over five independent runs with different random seeds. In each run, 5-fold cross-validation is performed on a 80\% training and 20\% test split. The best parameters were selected according to the model's accuracy on the validation set (10\% of the training set) across all 5 folds. We implemented early stopping with a patience of 80 epochs and a minimum difference in validation accuracy of $10^{-7}$. The validation set is used for both parameter tuning and early stopping.

\section{Results}\label{sec:results}

\begin{table*}[t]
%\small
\centering
\begin{tabular}{l|r@{\hspace{1\tabcolsep}} r@{\hspace{1\tabcolsep}} r@{\hspace{1\tabcolsep}}|r@{\hspace{1\tabcolsep}} r@{\hspace{1\tabcolsep}} r@{\hspace{1\tabcolsep}}|r@{\hspace{1\tabcolsep}} r@{\hspace{1\tabcolsep}} r@{\hspace{1\tabcolsep}}}
\toprule
 & \multicolumn{3}{c|}{Randomly initialized} & \multicolumn{3}{c|}{GloVe} & \multicolumn{3}{c}{BERT} \\ 
\textbf{Model} & \textbf{P} & \textbf{R} & \textbf{F\textsubscript{1} (std)} & \textbf{P} & \textbf{R} & \textbf{F\textsubscript{1} (std)} & \textbf{P} & \textbf{R} & \textbf{F\textsubscript{1} (std)} \\ \midrule
Baseline & 0.572 & 0.573 & 0.552 (0.07) & 0.751 & 0.738 & 0.728 (0.08) & 0.900 & 0.899 & 0.893 (0.04) \\ \midrule
+ noise & 0.599 & 0.574 & 0.541 (0.08) & 0.721 & 0.715 & 0.709 (0.09) & 0.914	& 0.916	& 0.913	(0.03) \\ \midrule
+ ET & \textbf{0.615} & \textbf{0.605} & \textbf{0.586} (0.06) & \textbf{0.795} & \textbf{0.786} & 0.781 (0.06) & 0.913 & 0.907 & 0.904 (0.05) \\ \midrule
+ EEG full &  0.540	& 0.538 &	0.525 (0.06) & 	0.738	& 0.729	& 0.725	(0.07)	&	0.913 &	0.909 &	0.906	(0.04)\\
+ EEG \texttheta & 0.602 &	0.599 &	0.584*	(0.08)	&	0.789 &	0.785 &	\textbf{0.783}+	(0.05)	&	\textbf{0.917} &	0.916 &	0.913*	(0.04) \\
+ EEG \textalpha & 0.610 &	0.590 &	0.565 	(0.05)	&	0.763 &	0.758 &	0.753 (0.05) &	0.912	& 0.908 &	0.906	(0.03)	\\
+ EEG \textbeta & 0.587	& 0.578	& 0.555	(0.07)	&	0.781 &	0.777 &	0.774+	(0.06)	&	0.911 &	0.911 &	0.907*	(0.04) \\
+ EEG \textgamma &  0.614 &	0.591 &	0.553	(0.08)	&	0.777 &	0.773 &	0.769*	(0.07)	&	\textbf{0.917}	& \textbf{0.917} &	\textbf{0.915}*	(0.04)\\
+\texttheta+\textalpha+\textbeta+\textgamma &  0.597 &	0.597 &	0.569	(0.08)	&	0.766 &	0.764 &	0.760*	(0.07)	&	0.913 &	0.913 &	0.911*	(0.04)\\ \bottomrule
\end{tabular}
\caption{\textbf{\textit{Binary sentiment analysis} results of the multi-modal model using the \textit{recurrent EEG decoding component}}. We report precision (P), recall (R), F\textsubscript{1}-score and the standard deviation (std) between five runs. The best results per column are marked in bold
%, all EEG results better than the text baseline \textit{and} the baseline augmented with random noise are marked with grey background
. Significance is indicated on the F\textsubscript{1}-score with asterisks: * denotes $p<0.05$ (uncorrected), 
+ denotes $p<0.003$ (Bonferroni corrected p-value).}
\label{tab:senti2-results}
\end{table*}

\begin{table}[t]
\centering
%\small
\begin{tabular}{l|c@{\hspace{1\tabcolsep}} c@{\hspace{1\tabcolsep}} c@{\hspace{1\tabcolsep}}|c@{\hspace{1\tabcolsep}} c@{\hspace{1\tabcolsep}} c@{\hspace{1\tabcolsep}}|c@{\hspace{1\tabcolsep}} c@{\hspace{1\tabcolsep}} c@{\hspace{1\tabcolsep}}}
\toprule
 & \multicolumn{3}{c|}{Randomly initialized} & \multicolumn{3}{c|}{GloVe} & \multicolumn{3}{c}{BERT} \\ 
\textbf{Model} & \textbf{P} & \textbf{R} & \textbf{F\textsubscript{1} (std)} & \textbf{P} & \textbf{R} & \textbf{F\textsubscript{1} (std)} & \textbf{P} & \textbf{R} & \textbf{F\textsubscript{1} (std)} \\ \midrule
Baseline & 0.408 & 0.384 &	0.351	(0.07)	&0.510	& 0.507 & 0.496	(0.06) & 0.722 & 0.714 & 0.710 (0.05) \\ \midrule
+ noise & 0.373 & 0.399 & 0.344 (0.10) & 0.531 & 0.519 & 0.504 (0.04) & 0.711 & 0.706 & 0.700 (0.06) \\ \midrule
+ ET & \textbf{0.424} & \textbf{0.413} & \textbf{0.388} (0.06) & \textbf{0.539} & \textbf{0.528} & \textbf{0.513} (0.04) & 0.728 & 0.717 & 0.714 (0.05) \\ \midrule
+ EEG full & 0.391	& 0.387 &	0.353	(0.07)	&	0.505 &	0.505	 & 0.488	(0.07) & 0.724 & 0.715 & 0.711 (0.06) \\
+ EEG \texttheta & 0.397	& 0.409	 & 0.360	(0.07)	&	0.516	& 0.510 &	0.498	(0.06) & 0.715 & 0.708 & 0.704 (0.05) \\
+ EEG \textalpha & 0.390 &	0.390 &	0.347	(0.08)	&	0.520 &	0.516 &	0.506	(0.05) & 0.720 & 0.712 & 0.707 (0.05) \\
+ EEG \textbeta & 0.350	& 0.370	& 0.302	(0.09)	&	0.523 &	0.519 &	0.509	(0.05) & \textbf{0.732} & \textbf{0.720} & \textbf{0.717} (0.07) \\
+ EEG \textgamma & 0.409 &	0.397 &	0.359	(0.07)	&	0.517. &	0.513 &	0.502	(0.04) & 0.709 & 0.705 & 0.697 (0.06) \\
+\texttheta+\textalpha+\textbeta+\textgamma & 0.401	& 0.400	& 0.368	(0.06)	&	0.522	& 0.516	& 0.505	(0.05) & 0.722 & 0.717 & 0.713 (0.05) \\ \bottomrule
\end{tabular}
\caption{\textbf{\textit{Ternary sentiment analysis} results of the multi-modal model using the \textit{recurrent EEG decoding component}}. We report precision (P), recall (R), F\textsubscript{1}-score and the standard deviation (std) between five runs. The best results per column are marked in bold.}
\label{tab:senti3-results}
\end{table}

\begin{table}[t]
\centering
%\small
\begin{tabular}{l|c@{\hspace{1\tabcolsep}} c@{\hspace{1\tabcolsep}} c@{\hspace{1\tabcolsep}}|c@{\hspace{1\tabcolsep}} c@{\hspace{1\tabcolsep}} c@{\hspace{1\tabcolsep}}|c@{\hspace{1\tabcolsep}} c@{\hspace{1\tabcolsep}} c@{\hspace{1\tabcolsep}}}
\toprule
 & \multicolumn{3}{c|}{Randomly initialized} & \multicolumn{3}{c|}{GloVe} & \multicolumn{3}{c}{BERT} \\ 
\textbf{Model} & \textbf{P} & \textbf{R} & \textbf{F\textsubscript{1} (std)} & \textbf{P} & \textbf{R} & \textbf{F\textsubscript{1} (std)} & \textbf{P} & \textbf{R} & \textbf{F\textsubscript{1} (std)} \\ \midrule
Baseline & 0.404 &	\textbf{0.525} &	\textbf{0.452}	(0.04)	&	0.501 &	0.609 &	0.539	(0.05)	&	0.522 &	\textbf{0.788} &	0.623	(0.05) \\ \midrule
+ noise & 0.420	& 0.424	& 0.408	(0.07) & 0.577 & 0.497 & 0.532 (0.03) & \textbf{0.675} & 0.585 & 0.625 (0.03) \\ \midrule
+ ET & 0.421	& 0.404	& 0.402	(0.06) & 0.547 & 0.476 & 0.506 (0.04) & 0.661 & 0.631 & 0.644 (0.03) \\ \midrule
+ EEG full &  0.345	& 0.343	& 0.334	(0.05)	&	0.511 &	0.387 &	0.432	(0.09)	&	0.652 &	0.690 &	0.668*	(0.10) \\
+ EEG \texttheta & \textbf{0.430} &	0.421 &	0.414	(0.07)	&	\textbf{0.582} &	0.508 &	0.539	(0.07)	&	0.646 &	0.736	& 0.684*	(0.08) \\
+ EEG \textalpha &  0.368	& 0.373	& 0.358	(0.12)	&	\textbf{0.582} &	0.515 &	\textbf{0.542}	(0.06)	&	0.652 &	0.715 &	0.679*	(0.07)\\
+ EEG \textbeta & 0.349	& 0.340	& 0.329	(0.09)	&	0.581 &	0.497 &	0.532	(0.10)	&	0.674 &	0.726 &	\textbf{0.696}+	(0.06)\\
+ EEG \textgamma & 0.410	 & 0.399 &	0.397	(0.05)	&	0.554 &	0.488 &	0.514	(0.09)	&	0.666. &	0.715 &	0.686*	(0.07) \\
+ \texttheta+\textalpha+\textbeta+\textgamma & 0.370 &	0.376 &	0.363 	(0.09)	&	0.554 &	0.488 &	0.514	(0.09)	&	0.675 &	0.646 &	0.659	(0.04)\\ \bottomrule
\end{tabular}
\caption{\textbf{\textit{Relation detection} results of the multi-modal model using the \textit{recurrent EEG decoding component}}. We report precision (P), recall (R), F\textsubscript{1}-score and the standard deviation (std) between five runs. The best results per column are marked in bold. Significance is indicated on the F\textsubscript{1}-score with asterisks: * denotes $p<0.05$ (uncorrected), + denotes $p<0.003$ (Bonferroni corrected p-value).}
\label{tab:reldet-results}
\end{table}

\begin{table*}[t]
%\small
\centering
\begin{tabular}{l|c@{\hspace{1\tabcolsep}} c@{\hspace{1\tabcolsep}} c@{\hspace{1\tabcolsep}}|c@{\hspace{1\tabcolsep}} c@{\hspace{1\tabcolsep}} c@{\hspace{1\tabcolsep}}|c@{\hspace{1\tabcolsep}} c@{\hspace{1\tabcolsep}} c@{\hspace{1\tabcolsep}}}
\toprule
 & \multicolumn{3}{c|}{Randomly initialized} & \multicolumn{3}{c|}{GloVe} & \multicolumn{3}{c}{BERT} \\ 
\textbf{Model} & \textbf{P} & \textbf{R} & \textbf{F\textsubscript{1} (std)} & \textbf{P} & \textbf{R} & \textbf{F\textsubscript{1} (std)} & \textbf{P} & \textbf{R} & \textbf{F\textsubscript{1} (std)} \\ \midrule
Baseline & 0.572 & 0.573 & 0.552 (0.07) & 0.751 & 0.738 & 0.728 (0.08) & 0.900 & 0.899 & 0.893 (0.04) \\ \midrule
+ noise & 0.558 & 0.584 & 0.528	(0.11) & 0.780 & 0.767	 & 0.762 (0.06) & 0.895 &	0.887 &	0.883	(0.05) \\ \midrule
+ ET & 0.617 & 0.623 & 0.610 (0.07) & 0.790 &	0.790 &	0.783	(0.06) & 0.896 & 0.887 & 0.881 (0.05) \\ \midrule
+ EEG full &  0.588	& 0.583	& 0.572	(0.04)	&	0.778 &	0.774 &	0.772+	(0.05)	&	\textbf{0.928}	 & \textbf{0.927} & \textbf{0.926}*	0.03 \\
+ EEG \texttheta & 0.564	& 0.569	& 0.535	(0.08)	&	\textbf{0.805}	& 0.792	& 0.791+	(0.04)	&	0.922 &	0.919	& 0.917*	(0.03) \\
+ EEG \textalpha & 0.596 &	0.593 &	0.563	(0.08)	&	0.775 &	0.781 &	0.772*	(0.08)	&	0.920 &	0.917 &	0.916*	(0.03) \\
+ EEG \textbeta & 0.605	& 0.597	& 0.580	(0.08)	&	0.802	& \textbf{0.797}	& \textbf{0.792}+	(0.05)	&	0.920 &	0.914 &	0.914*	(0.04)\\
+ EEG \textgamma & \textbf{0.640} &	\textbf{0.625} &	\textbf{0.611}+	(0.09) &	0.787 &	0.780 &	0.776+	(0.05)	&	0.905 &	0.905 &	0.901	(0.04) \\
+\texttheta+\textalpha+\textbeta+\textgamma &  0.599 &	0.579 &	0.558 	(0.07) &	0.800 &	0.794 &	0.786+	(0.05)	&	0.909 &	0.910 &	0.907	(0.04) \\ \bottomrule
\end{tabular}
\caption{\textbf{\textit{Binary sentiment analysis} results of the multi-modal model using the \textit{convolutional EEG decoding component}}. We report precision (P), recall (R), F\textsubscript{1}-score and the standard deviation (std) between five runs. The best results per column are marked in bold. Significance is indicated on the F\textsubscript{1}-score with asterisks: * denotes $p<0.05$ (uncorrected), + denotes $p<0.003$ (Bonferroni corrected p-value).}
\label{tab:senti2-results-cnn}
\end{table*}

\begin{table}[t]
\centering
%\small
\begin{tabular}{l|c@{\hspace{1\tabcolsep}} c@{\hspace{1\tabcolsep}} c@{\hspace{1\tabcolsep}}|c@{\hspace{1\tabcolsep}} c@{\hspace{1\tabcolsep}} c@{\hspace{1\tabcolsep}}|c@{\hspace{1\tabcolsep}} c@{\hspace{1\tabcolsep}} c@{\hspace{1\tabcolsep}}}
\toprule
 & \multicolumn{3}{c|}{Randomly initialized} & \multicolumn{3}{c|}{GloVe} & \multicolumn{3}{c}{BERT} \\ 
\textbf{Model} & \textbf{P} & \textbf{R} & \textbf{F\textsubscript{1} (std)} & \textbf{P} & \textbf{R} & \textbf{F\textsubscript{1} (std)} & \textbf{P} & \textbf{R} & \textbf{F\textsubscript{1} (std)} \\ \midrule
Baseline & 0.408	& 0.384 &	0.351	(0.07)	&	0.510	& 0.507	& 0.496	(0.06) & 0.722 & 0.714 & 0.710 (0.05) \\ \midrule
+ noise & 0.359	& 0.388	& 0.334	(0.09)	& 0.494	& 0.484	& 0.476	(0.07) & 0.715	 & 0.683 & 0.684	(0.05) \\ \midrule
+ ET & 0.417 &	0.399 &	0.372 	(0.05)	&	0.509	 & 0.512	& 0.500	(0.07) & 0.721 & 0.687 & 0.670 (0.05) \\ \midrule
+ EEG full &  0.365	& 0.384	& 0.333	(0.08)	&	0.488	& 0.484	& 0.476	(0.06) & 0.738 & 0.724 & \textbf{0.723}+ (0.04)\\
+ EEG \texttheta & 0.389	& 0.372	& 0.330	(0.06)	&	0.511	& 0.495	& 0.477	(0.06) & 0.727 & 0.718 & 0.716+ (0.05)\\
+ EEG \textalpha & 0.357	& 0.382	& 0.331	(0.11)	&	0.534 &	0.525 &	0.515+	(0.06) & 0.732 & 0.715 & 0.713+ (0.04)\\
+ EEG \textbeta & \textbf{0.425} &	\textbf{0.418} &	\textbf{0.378}	(0.08)	&	0.534 &	\textbf{0.529} &	\textbf{0.520}+	(0.05) & 0.727 & 0.717 & 0.715	(0.04)\\
+ EEG \textgamma & 0.404	& 0.406	& 0.360	(0.08)	&	\textbf{0.539}	& 0.521	& 0.514	(0.06) & \textbf{0.733} & \textbf{0.725} & 0.721+ (0.04)\\
+\texttheta+\textalpha+\textbeta+\textgamma &  0.384 &	0.402	& 0.354	(0.10)	&	0.517	& 0.504	& 0.488	(0.05) & \textbf{0.733} & 0.717 & 0.715 (0.06)\\ \bottomrule
\end{tabular}
\caption{\textbf{\textit{Ternary sentiment analysis} results of the multi-modal model using the \textit{convolutional EEG decoding component}}. We report precision (P), recall (R), F\textsubscript{1}-score and the standard deviation (std) between five runs. The best results per column are marked in bold. Significance is indicated on the F\textsubscript{1}-score with asterisks: * denotes $p<0.05$ (uncorrected), + denotes $p<0.003$ (Bonferroni corrected p-value).}
\label{tab:senti3-results-cnn}
\end{table}

\begin{table}[t]
\centering
%\small
\begin{tabular}{l|c@{\hspace{1\tabcolsep}} c@{\hspace{1\tabcolsep}} c@{\hspace{1\tabcolsep}}|c@{\hspace{1\tabcolsep}} c@{\hspace{1\tabcolsep}} c@{\hspace{1\tabcolsep}}|c@{\hspace{1\tabcolsep}} c@{\hspace{1\tabcolsep}} c@{\hspace{1\tabcolsep}}}
\toprule
 & \multicolumn{3}{c|}{Randomly initialized} & \multicolumn{3}{c|}{GloVe} & \multicolumn{3}{c}{BERT} \\ 
\textbf{Model} & \textbf{P} & \textbf{R} & \textbf{F\textsubscript{1} (std)} & \textbf{P} & \textbf{R} & \textbf{F\textsubscript{1} (std)} & \textbf{P} & \textbf{R} & \textbf{F\textsubscript{1} (std)} \\ \midrule
Baseline & 0.404 &	\textbf{0.525} &	\textbf{0.452}	(0.04)	&	0.501 &\textbf{0.609} &	0.539	(0.05)	&	0.522 &	\textbf{0.788} &	0.623	(0.05) \\ \midrule
+ noise & 0.424	& 0.299	& 0.342	(0.06)	&	0.547	& 0.441 &	0.486	(0.06)	&	0.532 &	0.493 &	0.511 (0.07) \\ \midrule
+ ET & 0.415 &	0.307 &	0.345	(0.08)	&	0.447 &	0.413. &	0.428	(0.07)	&	0.558 &	0.665 &	0.593	(0.13) \\ \midrule
+ EEG full & 0.225	& 0.225	& 0.225	(0.06)	&	0.548 &	0.408 &	0.464	(0.07)	&	0.647 &	0.664 &	0.650	(0.09) \\
+ EEG \texttheta &  \textbf{0.437}	& 0.380	& 0.400	(0.05)	&	0.620 &	0.493 &	0.547	(0.05)	&	\textbf{0.721} &	0.698 &	\textbf{0.707}+	(0.03)\\
+ EEG \textalpha &  0.372 &	0.366 &	0.352	(0.12) &	0.509 &	0.433 &	0.461	(0.12)	&	0.661 &	0.697 &	0.675+	(0.08)\\
+ EEG \textbeta & 0.394	& 0.328 &	0.338	(0.09)	&	0.627 &	0.479 &	0.541	(0.05)	&	0.643 &	0.646 &	0.640	(0.11) \\
+ EEG \textgamma &  0.405 &	0.363 &	0.366	(0.09)	&	\textbf{0.646} &	0.490 &	\textbf{0.555}	(0.04)	&	0.667 &	0.699 &	0.679+	(0.06) \\
+ \texttheta+\textalpha+\textbeta+\textgamma & 0.324 &	0.227 &	0.257	(0.11)	&	0.460 &	0.436 &	0.437 (0.14)	&	0.610 &	0.562 &	0.584	(0.05)\\ \bottomrule
\end{tabular}
\caption{\textbf{\textit{Relation detection} results of the multi-modal model using the \textit{convolutional EEG decoding component}}. We report precision (P), recall (R), F\textsubscript{1}-score and the standard deviation (std) between five runs. The best results per column are marked in bold. Significance is indicated on the F\textsubscript{1}-score with asterisks: * denotes $p<0.05$ (uncorrected), 
+ denotes $p<0.003$ (Bonferroni corrected p-value).}
\label{tab:reldet-results-cnn}
\end{table}

In this study, we assess the potential of EEG brain activity data to enhance NLP tasks in a multi-modal architecture. We present the results of all augmented models compared to the baseline results. As described above, we select the hyper-parameters based on the best validation accuracy achieved for each setting. 

The performance of our models is evaluated based on the comparison between the predicted labels (i.e., positive, neutral or negative sentiment for a sentence; or the relation type(s) in a sentence) and the true labels of the test set resulting in the number of \textit{true positives} (TP), \textit{true negatives} (TN), \textit{false positives} (FP), and \textit{false negatives} (FN) across the classified samples. The terms \textit{positive} and \textit{negative} refer to the classifier's prediction, and the terms \textit{true} and \textit{false} refer to whether that prediction corresponds to the ground truth label. The following decoding performance metrics were computed:

\noindent Precision is the fraction of relevant instances among the retrieved instances, and is defined as 
\begin{align}
  Precision = \frac{TP}{TP + FP} 
\end{align}

\noindent Recall is the fraction of the relevant instances that are successfully retrieved:
\begin{align}
  Recall = \frac{TP}{TP + FN} 
\end{align}

\noindent The F\textsubscript{1}-score is the harmonic mean combining precision and recall:
\begin{align}
  F_1 \, score = 2 \cdot \frac{Precision \cdot Recall}{Precision + Recall} 
\end{align}

\noindent For analyzing the results, we report macro-averaged precision (P), recall (R), and F\textsubscript{1}-score, i.e., the metrics are calculated for each label to counteract the label imbalance in the datasets.

The results for the multi-modal architecture using the \textit{recurrent} EEG decoding component are presented in Table \ref{tab:senti2-results} for binary sentiment analysis, Table \ref{tab:senti3-results} for ternary sentiment analysis, and Table \ref{tab:reldet-results} for relation detection. The first three rows in each table represent the uni-modal text baseline, the multi-modal noise and eye-tracking baselines. This is followed by the multi-modal models augmented with the full broadband EEG signals and each of the four frequency bands. Finally, in the last row, we also present the results of a multi-modal model with five components, where text and each frequency band extractors are learned separately and concatenated at the end.
In both sentiment tasks, the EEG data yields modest but consistent improvements over the text baseline for all word embeddings types. However, in the case of relation detection, the addition of either eye tracking or brain activity data is not helpful for randomly initialized embeddings and only beneficial in some settings using GloVe embeddings. Nevertheless, the combination of BERT embeddings and EEG data does improve the relation detection models. Generally, the results show a decreasing maximal performance per task with increasing task complexity measured in terms of the number of classes (see Section \ref{sec:task-compl} for a detailed analysis). 

Furthermore, the results for the multi-modal architecture using the \textit{convolutional} EEG decoding component are presented in Table \ref{tab:senti2-results-cnn} for binary sentiment analysis, Table \ref{tab:senti3-results-cnn} for ternary sentiment analysis, and Table \ref{tab:reldet-results-cnn} for relation detection. The results of this model architecture yield higher overall results, whereas the trend across tasks is similar to the models using the recurrent EEG decoding component, i.e., considerable improvements for both sentiment analysis tasks, but for relation detection the most notable improvements are achieved with the BERT embeddings. This validates the popular choice of convolutional neural networks for EEG classification tasks \citep{craik2019deep,schirrmeister2017deep}. While recurrent neural networks are often used in NLP and linguistic modelling (due to the left-to-right processing mechanism), CNNs have shown better performance at learning feature weights from noisy data (e.g., \citealp{kvist2019comparative}). Hence, our convolutional EEG decoding component is able to better extract the task-relevant linguistic processing information from the input data.

To assess the results, we perform statistical significance testing with respect to the text baseline in a bootstrap test as described in \citet{dror2018hitchhiker} over the F\textsubscript{1}-scores of the five runs of all tasks. We compare the results of the multi-modal models using text and EEG data to the uni-modal text baseline. In addition, we apply the Bonferroni correction to counteract the problem of multiple comparisons. We choose this conservative correction because of the dependencies between the datasets used \citep{dror2017replicability}.
Under the Bonferroni correction, the global null hypothesis is rejected if $p < \alpha/N$, where $N$ is the number of hypotheses \citep{bonferroni1936teoria}. In our setting, $\alpha = 0.05$
and $N = 18$, accounting for the combination
of the 3 embedding types and 6 EEG feature sets, namely broadband EEG; \texttheta, \textalpha, \textbeta \: and \textgamma \: frequency bands; and all four frequency bands jointly. For instance, in Table \ref{tab:senti2-results-cnn} the improvements in 6 configurations out of 18 are also still statistically significant under the Bonferroni correction (i.e., $p<0.003$), showing that EEG signals bring significant improvements in the sentiment analysis task. In the results tables, we mark significant results under both the uncorrected and the Bonferroni corrected p-value.

\section{Discussion}\label{sec:discussion}

The results show consistent improvements on both sentiment analysis tasks, whereas the benefits of using EEG data are only visible in specific settings for the relation detection task. EEG performs better than, or at least comparable to, eye tracking in many scenarios. This study shows the potential of decoding EEG for NLP and provides a good basis for future studies. Despite the limited
amount of data, these results suggest that augmenting NLP systems with EEG features is a generalizable approach.

In the following sections, we discuss these results from different angles. We contrast the performance of different EEG features, we compare the EEG results to the text baseline and multi-modal baselines (as described in Section \ref{sec:aug-base}), and we analyze the effect of different word embedding types. Additionally, we explore the impact of varying training set sizes in a data ablation study. Finally, we investigate the possible reasons for the decrease in performance for the relation detection task, which we associate with the task complexity. We run all analyses with both the recurrent \textit{and} the convolutional EEG components.

\subsection{EEG Feature Analysis}\label{sec:disc-features}

\noindent We start by investigating the impact of the various EEG features included in our multi-modal models. Different neurocognitive aspects of language processing during reading are associated with brain oscillations at various frequencies. We first give a short overview of the cognitive functions related to EEG frequency bands that are found in literature before discussing the insights of our results.

\textit{Theta} activity reflects cognitive control and working memory \citep{williams2019thinking}, and increases when processing semantic anomalies \citep{prystauka2019power}. Moreover, \citet{bastiaansen2002event} showed a frequency-specific increase in theta power as a sentence unfolds, possibly related to the formation of an episodic memory trace, or to incremental verbal working memory load. High theta power is also prominent during the effective semantic processing of language \citep{bastiaansen2005theta}. \textit{Alpha} activity has been related to attentiveness \citep{klimesch2012alpha}. Both theta and alpha ranges are sensitive to the lexical–semantic processes involved in language translation \citep{grabner2007event}. \textit{Beta} activity has been involved in higher-order linguistic functions such as the discrimination of word categories and the retrieval of action semantics as well as semantic memory, and syntactic processes, which support meaning construction during sentence processing. There is evidence that suggests that beta frequencies are important for linking past and present inputs and the detection of novelty of stimuli, which are essential processes for language perception as well as production \citep{weiss2012too}. Beta frequencies also affect decisions regarding relevance \citep{eugster2014predicting}. In reading, a stronger power-decrease in lower beta frequencies has been found for neutral compared to negative words \citep{scaltritti2020language}. Contrarily, emotional processing of pictures enhances \textit{gamma} band power \citep{muller1999processing}. Gamma-band activity has been used to detect emotions \citep{li2009emotion}, and increases during syntactic and semantic structure building \citep{prystauka2019power}. In the gamma frequency band, a power increase was observed during the processing of correct sentences in multiple languages, but this effect was absent following semantic violations \citep{penolazzi2009gamma,hald2006eeg}. Frequency band features have often been used in deep learning methods for decoding EEG in other domains, such as mental workload and sleep stage classification \citep{craik2019deep}.

The results show that our multi-modal models yield better results with filtered EEG frequency bands than using the broadband EEG signal on almost all tasks and embedding types, as well as on both EEG decoding components. Although all frequency band features show promising results on some embedding types and tasks (e.g., BERT embeddings and gamma features for binary sentiment analysis reported in Table \ref{tab:senti2-results}), the results show no clear sign of a single frequency band outperforming the others (neither across tasks for a fixed embedding type, nor for a fixed task and across all embedding types). For the sentiment analysis tasks, where both EEG decoding components achieve significant improvements, theta and beta features most often achieve the highest results. As described above, brain activity in each frequency band reflects specific cognitive functions. The positive results achieved using theta band EEG features might be explained by the importance of this frequency band for successful semantic processing. Theta power is expected to rise with increasing language processing activity \citep{kosch2020one}. Various studies have shown that theta oscillations are related to semantic memory retrieval and can be task-specific (e.g.,  \citealp{marko2019neural,giraud2012cortical,bastiaansen2005theta}). Overall, previous research shows how theta correlates with the cognitive processing involved in encoding and retrieving verbal stimuli (see \citet{kahana2006cognitive} for a review), which supports our results. The good performance of the beta EEG features might on one hand be explained by the effect of the emotional connotation of words on the beta response \citep{scaltritti2020language}. On the other hand, the role of beta oscillations in syntactic and semantic unification operations during language comprehension \citep{meyer2018neural,bastiaansen2006oscillatory} is also supportive of our results. 

Based on the complexity and extent of our results, it is unclear at this point whether a single frequency band is more informative for solving NLP tasks. Data-driven methods can help us to tease more information from the recordings by allowing us to test broader theories and task-specific language representations \citep{murphy2018decoding}, but our results also clearly show that restricting the EEG signal to a given frequency band is beneficial. More research is required in this area to specifically isolate the linguistic processing from the filtered EEG signals.

\subsection{Comparison to Multi-Modal Baselines}

\noindent The multi-modal EEG models often outperform the text baselines (at least for the sentiment analysis tasks). We now analyze how the EEG models compare to the two augmented baselines described in Section \ref{sec:aug-base} (i.e., eye tracking and models augmented with random noise). We find that EEG always performs better than or equal to the multi-modal text + eye tracking models. This shows how promising EEG is as a data source for multi-modal cognitive NLP. Although eye tracking requires less recording efforts, these results corroborate that EEG data contain more information about the cognitive processes occurring in the brain during language understanding.

As expected, the baselines augmented with random noise perform worse than the pure text baselines in all cases except for binary sentiment analysis with BERT embeddings. This model seems to deal exceptionally well with added noise. In the case of relation detection, when no improvement is achieved (e.g., for randomly initialized embeddings), the added noise harms the models similarly to adding EEG signals. It becomes clear for this task that adding the full broadband EEG features is worse than adding random noise (except with BERT embeddings), but some of the frequency band features clearly outperform the augmented noise baseline.

\subsection{Comparison of Embedding Types}

\noindent Our baseline results show that contextual embeddings outperform the non-contextual methods across all tasks. \citet{arora2020contextual} also compared randomly initialized, GloVe and BERT embeddings and found that with smaller training sets, the difference in performance between these three embedding types is larger. This is in accordance with our results, which show that the type of embedding has a large impact on the baseline performance on all three tasks. The improvements of adding EEG data in all three tasks are especially noteworthy when using BERT embeddings. In combination with the EEG data, these embeddings achieve improvements across all settings, including the full EEG broadband data as well as all individual and combined frequency bands. This shows that state-of-the-art contextualized word representations such as BERT are able to interact positively with human language processing data in a multi-modal learning scenario.

Augmenting our baseline with EEG data on the binary sentiment analysis tasks results in approximately +3\% F\textsubscript{1}-score across all the different embeddings with the recurrent EEG component. The gain is slightly lower at +1\% for all the embeddings in the ternary sentiment classification task. While there is no significant gain for relation detection with random and GloVe embeddings, the improvements with BERT embeddings reach up to +7\%. This shows that the improvements gained by adding EEG signals are not only dependent on the task, but also on the embedding type. In foresight, this finding might be useful in the future, when new embeddings will improve the baseline performance even further while possibly also increasing the gain from the EEG signals.

%- Relation detection has also a constant change (in that case decrease) of the performance at -5\% \\
%- More or less consistent increase or decrease in performance for a fixed task is interesting. Might be important for the future embeddings which will increase the baseline performance and possibly also the gain of using EEG features 

\begin{figure*}[t!]
	\centering
 \includegraphics[width=1\textwidth]{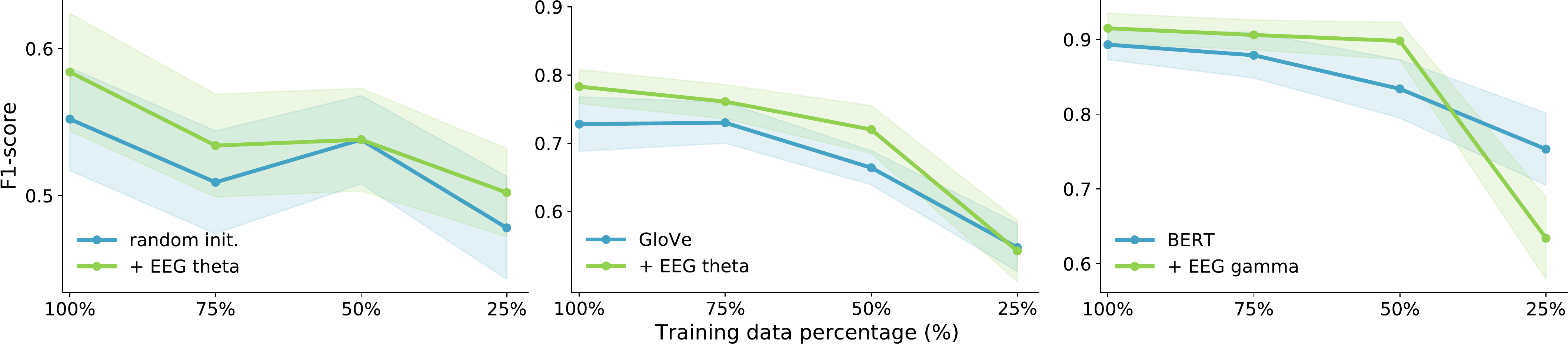}
 	\caption{Data ablation for all three word embedding types for the binary sentiment analysis task using the \textit{recurrent EEG decoding component}. The shaded areas represent the standard deviations.}
	\label{fig:data-abl-senti2-lstm}
\end{figure*}

\begin{figure*}[t!]
	\centering
 \includegraphics[width=1\textwidth]{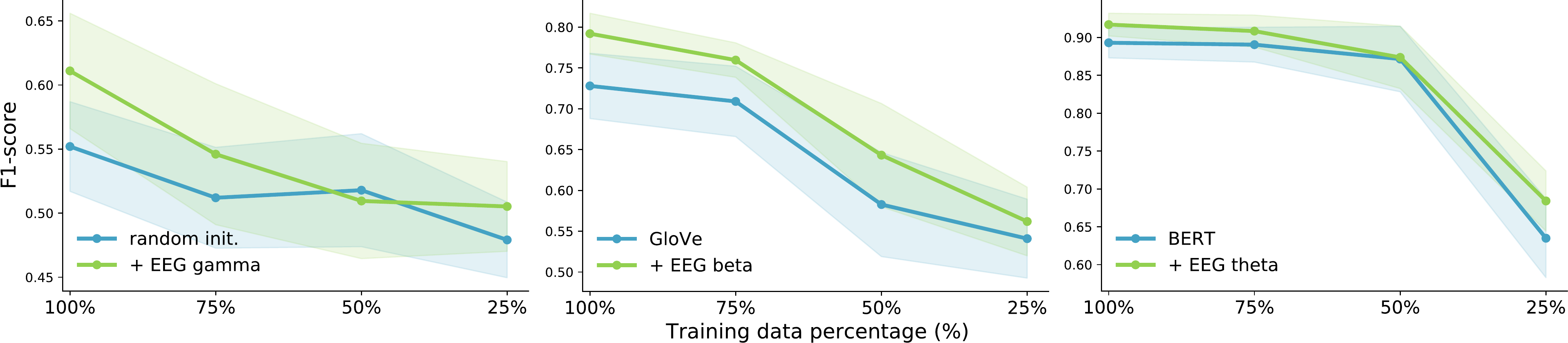}
 	\caption{Data ablation for all three word embedding types for the binary sentiment analysis task using the \textit{convolutional EEG decoding component}. The shaded areas represent the standard deviations.}
	\label{fig:data-abl-senti2-cnn}
\end{figure*}

\subsection{Data Ablation}

\noindent One of the challenges of NLP is to learn as much as possible from limited resources. Unlike most machine learning models, one of the most striking aspects of human learning is the ability to learn new words or concepts from limited numbers of examples \citep{lake2015human}.
Using cognitive language processing data may allow us to take a step towards meta-learning, the process of discovering the cognitive processes that are used to tackle a task in the human brain \citep{griffiths2019doing}, and in turn be able to improve the generalization abilities of NLP models. %Meta-learning aims to model the learning environment to make better use of limited data. 
Humans can learn from very few examples, while machines, particularly deep learning models, typically need many examples. Perhaps this advantage in humans is due to their multi-modal learning mechanisms \citep{linzen2020can}. 

Therefore, we analyze the impact of adding EEG features to our NLP models with less training data. We performed data ablation experiments for all three tasks. The most conclusive results were achieved on binary sentiment analysis. Randomly initialised embeddings unsurprisingly suffer a lot when reducing training data. The results are shown in Figure \ref{fig:data-abl-senti2-lstm} and \ref{fig:data-abl-senti2-cnn}, for both EEG decoding components. We present the results for the best-performing frequency bands only. The largest gain from EEG data is obtained with only 50\% of the training data with GloVe and BERT embeddings, which is as little as 105 training sentences. These experiments emphasize the potential of EEG signals for NLP especially when dealing with very small amounts of training data and using popular word embedding types.

\subsection{Task Complexity Ablation}\label{sec:task-compl}

\begin{table}[t]
\centering
%\small
\begin{tabular}{l|ccc|ccc}
\toprule
 & \multicolumn{3}{c|}{Recurrent EEG Decoding} & \multicolumn{3}{c}{Convolutional EEG Decoding} \\ \midrule
\textbf{Job Title vs. None} & \textbf{Precision} & \textbf{Recall} & \textbf{F\textsubscript{1}-score} & \textbf{Precision} & \textbf{Recall} & \textbf{F\textsubscript{1}-score}\\\midrule
GloVe & 0.789 & 0.776 & 0.767 (0.05) & 0.789 & 0.776 & 0.767 (0.05)\\
GloVe + EEG full & \textbf{0.792} & 0.782 & 0.773 (0.06) & 0.796 & 0.793 & 0.789 (0.05)\\
GloVe + EEG \textgamma & 0.780 & \textbf{0.788} & \textbf{0.774} (0.1) & \textbf{0.817} & \textbf{0.811} & \textbf{0.808} (0.03)\\\midrule
\textbf{Visited vs. None} & \textbf{Precision} & \textbf{Recall} & \textbf{F\textsubscript{1}-score} & \textbf{Precision} & \textbf{Recall} & \textbf{F\textsubscript{1}-score}\\\midrule
GloVe & 0.762 & 0.756 & 0.734 (0.1) & 0.762 & 0.756 & 0.734 (0.1) \\
GloVe + EEG full & 0.756 & 0.759 & 0.745 (0.1) & 0.766 & 0.758 & 0.750 (0.09) \\
GloVe + EEG \textgamma & \textbf{0.773} & \textbf{0.768} & \textbf{0.754} (0.1) &  \textbf{0.819} & \textbf{0.795} & \textbf{0.795} (0.09)\\\bottomrule
\end{tabular}
\caption{Binary relation detection results for both EEG decoding components for the relation types \textit{Job Title} and \textit{Visited} using GloVe embeddings. The best result in each column is marked in bold.}
\label{tab:bin-reldetect}
\end{table}

\noindent From the previously described results, one hypothesis on the reason why augmenting the baseline with EEG data lowers the performance in the relation detection task with randomly initialized and GloVe embeddings lies in the complexity of the task. More concretely, we measure the complexity by counting the number of classes the model needs to learn. Generally, more complex tasks (in terms of number of classes) require more data to generalize (see for instance \citealp{li2018multi}). Therefore, it is clear that with a fixed amount of data, the impact of augmenting the feature space with additional information (in this case EEG data) is also less visible for the more complex tasks. We see a decrease in performance with increasing complexity over the three evaluated tasks with all embeddings except for BERT. Therefore, we validate this hypothesis by simplifying the relation detection task by reducing the number of classes from 11 to 2. We create binary relation detection tasks for the two most frequent relation types \textit{Job Title} and \textit{Visited} (see Figure \ref{fig:deco-rel-distr}). For example, we classify all the samples containing the relation \textit{Job Title} (184 samples) against all samples with no relation (219 samples).

We train these additional models with GloVe embeddings, since these did not show any significant improvements when augmented with EEG data on the full relation detection task. The results for the full broadband EEG features and the best frequency band from the previous convolutional results (gamma) are shown in Table \ref{tab:bin-reldetect}. It is evident that with the simplification of the relation detection task into binary classification tasks, EEG signals are able to boost the performance of the non-contextualized Glove embeddings and achieve considerable improvements over the text baseline. The gains are similar as for binary sentiment analysis for both EEG decoding components. This confirms our hypothesis that the EEG features tested yield good results on simple tasks, but more research is needed to achieve improvements on more complex tasks. Note that, as mentioned previously, this is not the case for BERT embeddings, which outperform the baselines on all NLP tasks.

\subsection{Conclusion}

\noindent We presented a large-scale study about leveraging electrical brain activity signals during reading comprehension for augmenting machine learning models of semantic language understanding tasks, namely, sentiment analysis and relation detection. We analyzed the effects of different EEG features and compared the multi-modal models to multiple baselines. Moreover, we compared the improvements gained from the EEG signals on three different types of word embeddings. Not only did we test the effect of varying training set sizes, but also tasks of various difficulty levels (in terms of number of classes).

We achieve consistent improvements with EEG across all three embedding types. The models trained with BERT embeddings yield significant performance increases on all NLP tasks. However, for randomly initialized and GloVe embeddings the improvement magnitude decreases for more difficult tasks. For these two types of embedding, the improvement for the binary and ternary sentiment analysis tasks ranges between 1-4\% F\textsubscript{1}-score. For relation detection, a multi-class and multi-label sequence classification task, it was not possible to achieve any improvements unless the task complexity is substantially reduced. Therefore, our experiments show that state-of-the-art contextualized word embeddings combined with careful EEG feature selection achieve good results in multi-modal learning.
Moreover, we find that in the tasks where the multi-modal architecture does achieve considerable improvements, the convolutional EEG decoding component yields even higher results than the recurrent component.

To sum up, we capitalize on the advantages of electroencephalography data to
examine if and which EEG features can serve to augment language understanding models. While our results show that there is linguistic information in the EEG signal complementing the text features, more research is needed to isolate language-specific brain activity features. More generally, this work paves the way for more in-depth EEG-based NLP studies. 

\bibliographystyle{johd}
\bibliography{bib}

%\section*{Supplementary Files (optional)}
%Any supplementary/additional files that should link to the main publication must be listed, with a corresponding number, title and option description. Ideally the supplementary files are also cited in the main text.
%Note: supplementary files will not be typeset so they must be provided in their final form. They will be assigned a DOI and linked to from the publication.

\end{document}